*Article*

# Machine Learning Techniques for Estimating Soil Moisture from Smartphone Captured Images

**Muhammad Riaz Hasib Hossain [1,*] and Muhammad Ashad Kabir [1,2,]**

[1] School of Computing, Mathematics, and Engineering, Charles Sturt University, Bathurst, NSW 2795, Australia, akabir@csu.edu.au
[2] Gulbali Institute for Agriculture, Water and Environment, Charles Sturt University, Wagga Wagga, NSW 2678, Australia
* Correspondence: muhossain@csu.edu.au (M.R.H.H.)

**Abstract:** Precise Soil Moisture (SM) assessment is essential in agriculture. By understanding the level of SM, we can improve yield irrigation scheduling which significantly impacts food production and other needs of the global population. The advancements in smartphone technologies and computer vision have demonstrated a non-destructive nature of soil properties, including SM. The study aims to analyze the existing Machine Learning (ML) techniques for estimating SM from soil images and understand the moisture accuracy using different smartphones and various sunlight conditions. Therefore, 629 images of 38 soil samples were taken from seven areas in Sydney, Australia, and split into four datasets based on the image-capturing devices used (iPhone 6s and iPhone 11 Pro) and the lighting circumstances (direct and indirect sunlight). A comparison between Multiple Linear Regression (MLR), Support Vector Regression (SVR), and Convolutional Neural Network (CNN) was presented. MLR was performed with higher accuracy using holdout cross-validation, where the images were captured in indirect sunlight with the Mean Absolute Error (MAE) value of 0.35, Root Mean Square Error (RMSE) value of 0.15, and $R^2$ value of 0.60. Nevertheless, SVR was better with MAE, RMSE, and $R^2$ values of 0.05, 0.06, and 0.96 for 10-fold cross-validation and 0.22, 0.06, and 0.95 for leave-one-out cross-validation when images were captured in indirect sunlight. It demonstrates a smartphone camera's potential for predicting SM by utilizing ML. In the future, software developers can develop mobile applications based on the research findings for accurate, easy, and rapid SM estimation.

## 1. Introduction

Soil is a fundamental element for food production and other needs of the global population. Optimizing natural resources is a crucial aspect of supplying the growing population with food [1]. One significant element in cultivation is soil moisture, which regulates the flow of water and energy from the atmosphere to the field [2]. Soil moisture is the quantity of water in the soil expressed as a percentage. The moisture in the soil varies and changes with time. It depends upon several factors, such as the amount of rain in a particular area, irrigation, the consumption of liquid in soils via evaporation, and so on [3]. It maintains an enhanced relationship with climate change. Various temperatures, precipitation, and other climate factors lead to variations in soil moisture [4]. Therefore, farmers can use moisture conditions to determine the health and productivity of crops in order to maximize irrigation [2].

Traditionally, several moisture prediction methods are available. Farmers in developed countries often rely on a laboratory-based gravimetric analysis to predict soil moisture [5–8]. However, it is time-consuming, costly, and unavailable in many places. Various tools such as the SDI-12 Sensor Reader, Time Domain Reflectometry (TDR), Frequency Domain Reflectometry (FDR), and tensiometers are other alternatives for moisture estimation. Still, these tools are costly for farmers [8,9]. Recently, satellite remote sensing has been implemented to predict soil moisture in a broader area than in small farmlands [9]. However, this technique does not show higher accuracy for soil moisture estimation [9]. Considering the disadvantages of traditional methods, an effective and cheap non-destructive alternative for moisture estimation can be a soil color identification technique, as the color of the soil switches according to the changing moisture condition of the soil [5,7,10,11].

The color of the soil is a vital characteristic of soil identification [12]. Several techniques can be used for the comparison of soil colors. One popular technique is a Munsell colorimetric system that requires a visual correspondence between color chips and a soil sample [13]. However, the Munsell color chart is unsuitable for precise measurements of soil colors due to limited standard color chips [11,13]. Moreover, specific color chips of the Munsell color chart are hard to distinguish with the naked eye, which can impact the estimation [6,14]. This restriction can be resolved by another method, such as soil characterization using images from a digital camera or a built-in smartphone camera. It allows significant physical measurements of soil colors [7,11,15].

Machine Learning (ML) algorithms can provide predictions based on several features (for example, soil color values, moisture values, and so on). A feature is a measurable property of an object to be predicted and appears as a column within a dataset. ML has a mechanism to determine patterns and discover knowledge from datasets. The utilization of smartphone cameras, algorithms of ML, or techniques of artificial intelligence have demonstrated their ability to provide a rapid and non-destructive nature of soil elements. For this reason, a smartphone-based soil moisture measurement will be quicker, less costly, and easier to assess [11].

This paper examines the existing ML techniques to estimate soil moisture using smartphone-captured images through experimental research. We compared multiple ML models to discover the ideal model for moisture estimation from soil images. Moreover, a comparative analysis of the research results is done to understand the differences in moisture accuracy using various smartphone devices as well as direct and indirect sunlight conditions during soil image capture. The long-term goal of this experimental research is to develop a smartphone application based on the findings so that farmers can estimate the soil moisture of their lands quickly, cheaply, and efficiently in the future. Additionally, the collected data, including soil images, will be publicly available for future research. This paper has made the following two contributions:

- A comparison has been established among the ML models throughout the experiment research to find a better, more efficient model for soil moisture estimation.
- A comparative analysis of the research findings has been presented to understand the differences in moisture accuracy using various smartphone devices and direct and indirect sunlight conditions during soil image capture.

The remaining portions of the paper are broken down into subsequent sections. Section 2 presents related works, including the limitations of techniques or approaches of various ML models used previously for soil moisture estimation using soil images. Section 3 describes the materials and methodology. Section 4 documents the experimental setting of this study. Section 5 sets out the empirical findings and discussions about the outcome. The last section includes the conclusions containing achievements, future works, and a summary of this study's results.

## 2. Related Works

Several approaches and techniques written on predicting soil moisture based on soil images were reviewed. Several demerits were found in those approaches, which are also discussed.

### *2.1. Machine Learning Models for Soil Moisture Prediction*

Although ML was widely used for forecasting soil moisture from the numerical data collected using several types of equipment, such as a data logger, few researchers worked on ML models embedded with soil images. Table 1 lists the ML models for predicting soil moisture conditions from images captured with digital cameras or smartphones.

**Table 1.** List of ML models with descriptions that were used to predict soil moisture.

| No. | ML Model | Description |
|---|---|---|
| 01 | Artificial Neural Network (ANN) | An ANN model is a subset of machine learning and is the heart of deep learning. It consists of input, hidden, and output layers. In addition, it includes multiple connected processing units that work together to process information [16]. |
| 02 | Cubist | A Cubist model is an addition to Quinlan's M5 approach. Though it generates a tree, each path of the tree is reduced to a rule, and linear regression models are contained in the terminal nodes. In addition, rules are pruned or combined to simplify the model [17]. |
| 03 | Convolutional Neural Network (CNN) | A CNN is a subclass of an ANN. Input, hidden, and output layers make up its structure. It is used especially for image recognition [16]. |
| 04 | Gaussian Process Regression (GPR) | A GPR model is a kernel-based machine learning model used for accurate predictions [11]. |
| 05 | Linear Regression (LR) | An LR model displays the relation of two variables for prediction. A simple linear regression model implements an independent variable to predict a dependent variable. Nevertheless, multiple linear regression is a supervised ML algorithm with multiple independent variables and a single dependent variable for regression [18]. |
| 06 | Multilayer Perceptron (MLP) | An MLP network is an ANN that comprises a group of units with an input layer, one or more hidden layers, and a single output layer. Output activation in the computation nodes is generated by a nonlinear activation function named the sigmoid function. The model uses a backpropagation algorithm to train regressions [19]. |
| 07 | Partial Least Squares (PLS) | A PLS regression uses a set of independent predictors or variables to predict a group of dependent variables. It is handy when there are strong collinear predictors or more predictors than observations, and regression of Ordinary Least Squares (OLS) produces coefficients with high standard errors or fails [20]. |
| 08 | Support Vector Regression (SVR) | An SVR known as a Support Vector Machine (SVM) regression is applied to predict numeric values rather than classifications. It is a proficient prediction model that recognizes the existence of nonlinearity in the data. A straight line is required to fit the data in SVR and is called a hyperplane [21]. |
| 09 | Random Forest (RF) | An RF is a supervised ML algorithm accepted for classification and regression. It is constructed from decision tree algorithms that predict behavior and outcome [16]. |
| 10 | Regression Trees | Regression trees evaluate the association between dependent and independent variables [16]. |

Previously, an LR model was found to be more successful than other research algorithms. For example, the authors in [7] used LR for moisture prediction from soil images. They obtained satisfactory results from simple and multiple LR models based on soil classification. In [10], the authors also captured satisfactory results using an LR model and found that the soil moisture was high in light-colored soil. On the other hand, the authors in [22] achieved moderate accuracy (only 65%) with an LR model for soil moisture prediction.

Some research achieved satisfactory results with other models than LR. The authors in [8] observed that SVR made better predictions compared with MPL when trained with a single type of soil data. Although both models predicted soil moisture at high correlation coefficient ($R$) values, the range of $R$ in the SVR was 0.89 to 0.99. Furthermore, the authors in [5] found a satisfactory outcome (RMSE between 0.0321 and 0.0650 g/g and r2 between 0.6675 and 0.8231) with an ANN for soil moisture prediction when a hidden layer with twelve neurons and the tan-sigmoid transfer function, were used. In [6], the authors also noted that the backpropagation neural networks performed better than PLS in their research. On the other hand, the authors in [11] obtained superior performances using a GPR model and a Cubist model from past experimental research. Table 2 summarizes the literature and the limitations of ML models for predicting soil moisture from images.

**Table 2.** Overview of literature and limitations.

| Paper | Experimental Details | Limitation(s) |
|---|---|---|
| [6] | Model(s): ANN and PLS<br>Best Performances: ANN trained with RGB color space and site-specific data (land cover, vegetative cover, canopy cover, altitude, profile depth, slope, landform, and topography)<br>Soil Sample Size: 273 samples<br>Sample Collection: Halaba area of southwest Ethiopia | Although the paper indicated that grouping samples by soil type increased model performance, grouping samples according to the soil types was not done. |
| [8] | Model(s): SVR and MLP<br>Best Performances: SVR<br>Soil Sample Size: Thirty-five soil samples of six soil types<br>Sample Collection: Chateau Kefraya terroirs in Lebanon | Many other factors, such as the soil's physical, chemical, and biological components, could have been responsible for the soil color variation along with soil moisture. However, these properties were not evaluated in the research. |
| [10] | Model(s): Simple LR model<br>Best Performances: Satisfactory result was found using a simple LR model | Based on the limited data, the paper presented that soil color and soil moisture were strongly related. It also found that soils became darker when soil moisture |
| | Soil Sample Size: Five soils (four are natural soils, and one is fine sand) have up to twenty-seven samples for each soil<br>Sample Collection: Four places (Löddeköpinge, Värpinge, Lund, and Odarslöv) in Sweden | increased. However, some lighter soil colors indicated the highest soil moisture in the research. Further investigation was needed with extensive data. |
| [22] | Model(s): LR models<br>Best Performances: Moderate accuracy by LR<br>Soil Sample Size: Eight samples of Alfisol soil type<br>Sample Collection: Karanganyar District, Indonesia | Soil moisture estimation was moderately accurate (65%). Moreover, samples were collected from a single area, and the scope for samples from other geographical sectors was not considered. |
| [7] | Model(s): Simple LR model and Multiple Linear Regression (MLR) models<br>Best Performances: Simple LR model or MLR model based on soil types<br>Soil Sample Size: Six soil samples<br>Sample Collection: Federal University of Viçosa (UFV) | Soil moisture was predicted from the soil surface, which may differ from the inner soil sample. Another limitation was that soil characteristics must be analyzed before the model selection, which was not done. Moreover, the result may not be satisfactory for all soil classes because complementary studies were not conducted for different soil classes to predict soil moisture. |
| [11] | Model(s): 24 ML models (6 LR models, 4 GPR models, 3 Decision Tree models, 6 SVM, 4 Ensembles of Decision Tree models, and ANN)<br>Best Performances: GPR model and Cubist model<br>Soil Sample Size: Twenty-five samples from two agricultural fields<br>Sample Collection: MacDonald Campus Farm, McGill University, Quebec, Canada | High moisture content was found in the dark-colored soils. However, soil color may be related to soil type contrasts, textural differences, and other factors such as topography, geology, climate, and so on, which were not considered explicitly. |
| [5] | Model(s): ANNs<br>Best Performances: ANN with the tan-sigmoid transfer function and a hidden layer containing 12 neurons<br>Soil Sample Size: Three types of soil<br>Sample Collection: Alegre, Espírito Santo, Brazil; and Guaçuí, Espírito Santo, Brazil | No experiments were conducted for a more robust characterization of soil color variation to estimate the soil moisture content. |
| [23] | Model(s): ANNs<br>Best Performances: ANN with the tan-sigmoid transfer function and a hidden layer containing 12 neurons<br>Soil Sample Size: Three types of soil<br>Sample Collection: Alegre, Espírito Santo, Brazil; and Guaçuí, Espírito Santo, Brazil | No experiments were conducted for a more robust characterization of soil color variation to estimate the soil moisture content. |
| [9] | Model(s): MLR<br>Best Performances: MLR with G (Green), B (Blue), H (Hue), and S (Saturation) input parameters<br>Soil Sample Size: Samples from 40 test sites<br>Sample Collection: A farmland in Beijing, China | The research was done based on a single soil type. Therefore, heterogeneous soil types were not considered, which might present a different result. |

Many studies are being performed to predict soil moisture using various ML algorithms. While several models are already developed for predicting soil moisture, there is still a place to enhance the accuracy of a model with varied input parameters. Another concern is the high-quality data that is essential to form an ML model. Loud, dirty, and incomplete data are the unavoidable enemies of a perfect ML. Therefore, this paper proposes an ML algorithm after comparing multiple models trained by high-quality data (features) with various input parameters to estimate soil moisture.

## 2.2. Image Capturing Devices

Several researchers used digital cameras to capture soil images for moisture content estimation. For example, the authors in [5,8,9] used an 18 megapixels digital camera (Canon EOS 1200D), a 7.1 megapixels digital camera (Canon PowerShot A710 IS), and a 3.2 megapixels digital camera (Canon A310), respectively, for color photographs of soils. The authors in [7,13] also used digital cameras—a Nikon Coolpix L810 with a 4–104 mm lens and Nikon D100 with a 50-mm lens, respectively—to capture soil images. Similarly, the authors in [22] used a Samsung digital camera to capture soil images for their research.

Few researchers utilized smartphone cameras to capture soil images for soil moisture estimation. In this case, the authors in [11] used a smartphone device, an LG G5, and the authors in [6] used a Sony Xperia z3+ smartphone to capture sample images for soil moisture prediction.

Previously, multiple versions of any device were not utilized in research for capturing soil sample images. Since smartphone camera features are improving gradually, it is necessary to investigate the impact on soil moisture prediction. In this case, no research has been conducted to understand the effect of predicting soil moisture from images of soil samples taken with several smartphone models.

## 2.3. Lighting Conditions during Image Capture

The authors in [5,7,11,15,22] captured the images of soil samples at a laboratory using a fixed light. In [5], the authors used standardized light to avoid bias in the ANN model. The authors in [7,10] adjusted the white balance using a camera pre-setting and utilized fluorescent lamps for lighting before adopting images of the soil samples in a homogeneous light state in a laboratory. In [8], the authors implemented a continuous light source with white foam panels to ensure soft light in a dark room to capture the sample images. Many researchers rely on fixed light sources and distances, which allow for measuring the sensitive soil color from the image of the soil sample [24]. However, the authors in [9] took soil images from the field on sunny days instead of in a laboratory environment. The images were collected between 11 am and 2 pm to maintain relative light intensity. In [6], the authors also photographed the samples from the field under well-illuminated conditions.

Formerly, a fixed distance with a still flash was used in a laboratory to capture images of soil samples. The authors in [7,11, 22] captured the images from 23 cm, 25 cm, and 32 cm above the soil samples, respectively. In [8,10], the authors placed the camera approximately one meter and 0.5 meters above the table while capturing soil images. It is noted that numerous researchers did not take images from the field because soil moisture prediction is difficult without a laboratory environment [22]. However, the restricted laboratory condition differs from actual field conditions [11]. In this case, the authors in [9] captured the soil images from the field and maintained a distance of 100 meters.

Implementing the same distance and standardized lighting methodology is technically tricky, quite expensive, and time-consuming for farmers to achieve while capturing soil images from agricultural fields. Hence, research is needed to discover a better technique for farmers to take soil images directly from fields without a fixed distance to predict soil moisture. The authors in [25] tested if the soil color assessments were accurate when the smartphone camera took the images in sunny instead of cloudy conditions. However, the effects of direct and indirect sunlight conditions were not clarified. Therefore, it is necessary to experiment with sunlight's direct and indirect effects when an image is taken from the field.

We found that several works were done on soil moisture prediction using images. However, for a more accurate forecast, many studies used additional tools for collecting values of several features with soil images, which are expensive for farmers. Moreover, we did not find any research that used multiple devices to capture soil images needed to determine the effect on soil moisture estimation. They also captured the images at a fixed distance with constant lighting conditions rather than natural sunlight.

This paper proposes experimental research to further analyze the existing ML techniques for soil moisture prediction using smartphone images. This study used only one additional feature (parameter) for better moisture prediction. In this case, no additional tool or device was implemented to find the data of the additional feature because we utilized a smartphone app for this purpose. This study also focuses on a comparative analysis for understanding the differences in moisture prediction accuracy using various smartphone devices as well as direct and indirect sunlight conditions rather than fixed lighting during soil image capture. Furthermore, this research will assist in developing a smartphone application based on research results so that farmers can estimate soil moisture on their lands quickly, cheaply, and efficiently in the future.

## 3. Materials and Methodology

### *3.1. Soil Samples*

#### 3.1.1. Fields of Study

The sample data were taken in seven different landscape areas (Figure 1) in Sydney, NSW, Australia. Field investigations were conducted between 31 January 2022 and 16 March 2022 (Table 3). Selected landscape areas exhibited considerable spatial diversity of soil types. Therefore, various soil types were collected as part of the research for broader and stronger training models. There were thirty-eight soil samples collected from landscaped areas. Several instruments were employed for collecting soil sample information. A shovel was used to clean the surface before excavating a soil sample. This research used a soil sampler to extract an undisturbed soil profile. Around 20 cm depth of soil was collected using the soil sampler.

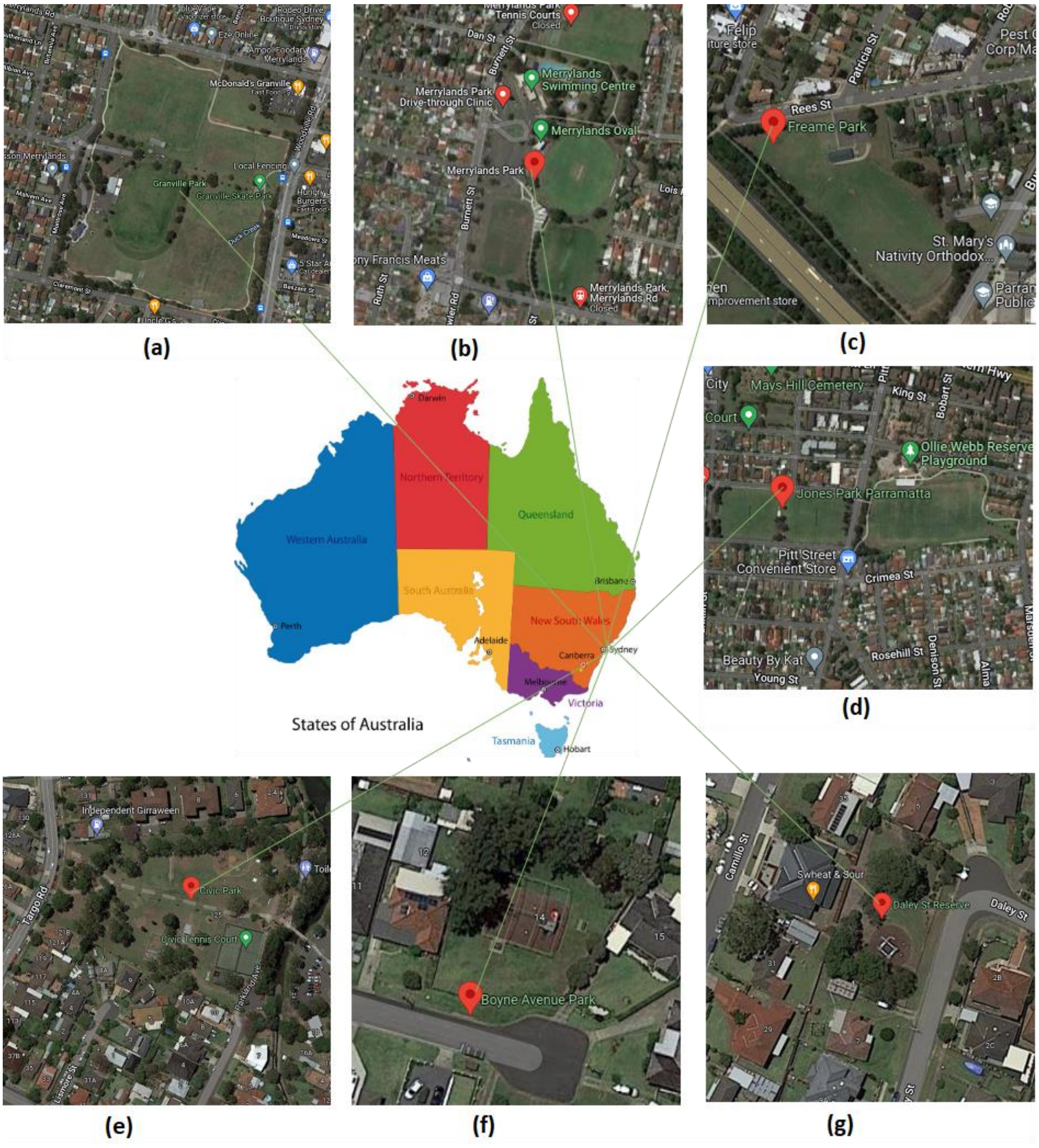

**Figure 1.** Geographical locations of sample collection points: (**a**) Granville Park, Merrylands, NSW, 2160; (**b**) Merrylands Park, Merrylands, NSW, 2160; (**c**) Freame Park, Mays Hill, NSW, 2145; (**d**) Jones Park Parramatta, Parramatta, NSW, 2145; (**e**) Civic Park, Pendle Hill, NSW, 2145; (**f**) Boyne Avenue Park, Pendle Hill, NSW, 2145; and (**g**) Daley St Reserve, Pendle Hill, NSW 2145.

**Table 3.** Total soil samples and harvest dates for each landscape area.

| No. | Area Name | Total Soil Samples | Collection Date |
|---|---|---|---|
| 01 | Granville Park, Merrylands, NSW, 2160 | | 31 January 2022 |
| 02 | Merrylands Park, Merrylands, NSW, 2160 | 09 | 02 February 2022 |
| 03 | Freame Park, Mays Hill, NSW, 2145 | 03 | 02 March 2022 |
| 04 | Jones Park Parramatta, Parramatta, NSW, 2145 | 07 | 02 March 2022 |
| 05 | Civic Park, Pendle Hill, NSW, 2145 | 05 | 02 March 2022 |
| 06 | Boyne Avenue Park, Pendle Hill, NSW, 2145 | 03 | 16 March 2022 |
| 07 | Daley St Reserve, Pendle Hill, NSW 2145 | 03 | 16 March 2022 |

### 3.1.2. Soil Analysis and Soil Imaging

To determine the moisture value of a soil sample, many researchers implemented a method known as gravimetric or thermogravimetric analysis. The soil samples are dried, crushed, sieved, and weighted in this method to measure the mass change. Usually, a laboratory oven was used to dry the samples until reaching a specific weight [5,7–11,22]. Though the gravimetric method is efficient in predicting soil moisture, it is expensive, and so is not widely used [5].

Several researchers used various devices to predict the moisture level of soil samples. For example, the authors in [5] employed a TDR moisture meter named 'TDR100', which has three parallel rod probes (CS-610) to estimate the soil's water content by evaluating the reflected waveform. A similar moisture meter device named 'SDI-12 Sensor Reader' (Figure 2) is used to collect soil moisture samples to avoid complexity and improve the actual moisture value for this research. This device was utilized for obtaining data, including temperature, permittivity, and moisture of the soil sample. Simultaneously, the latitude and longitude for each soil sample's location and collection time were documented with a Global Positioning System (GPS) integrated with a moisture meter. The moisture range of the collected thirty-eight soil samples was between 0.71% and 30.11%, and the standard deviation was 8.30, indicating soil moisture variability (Table 4).

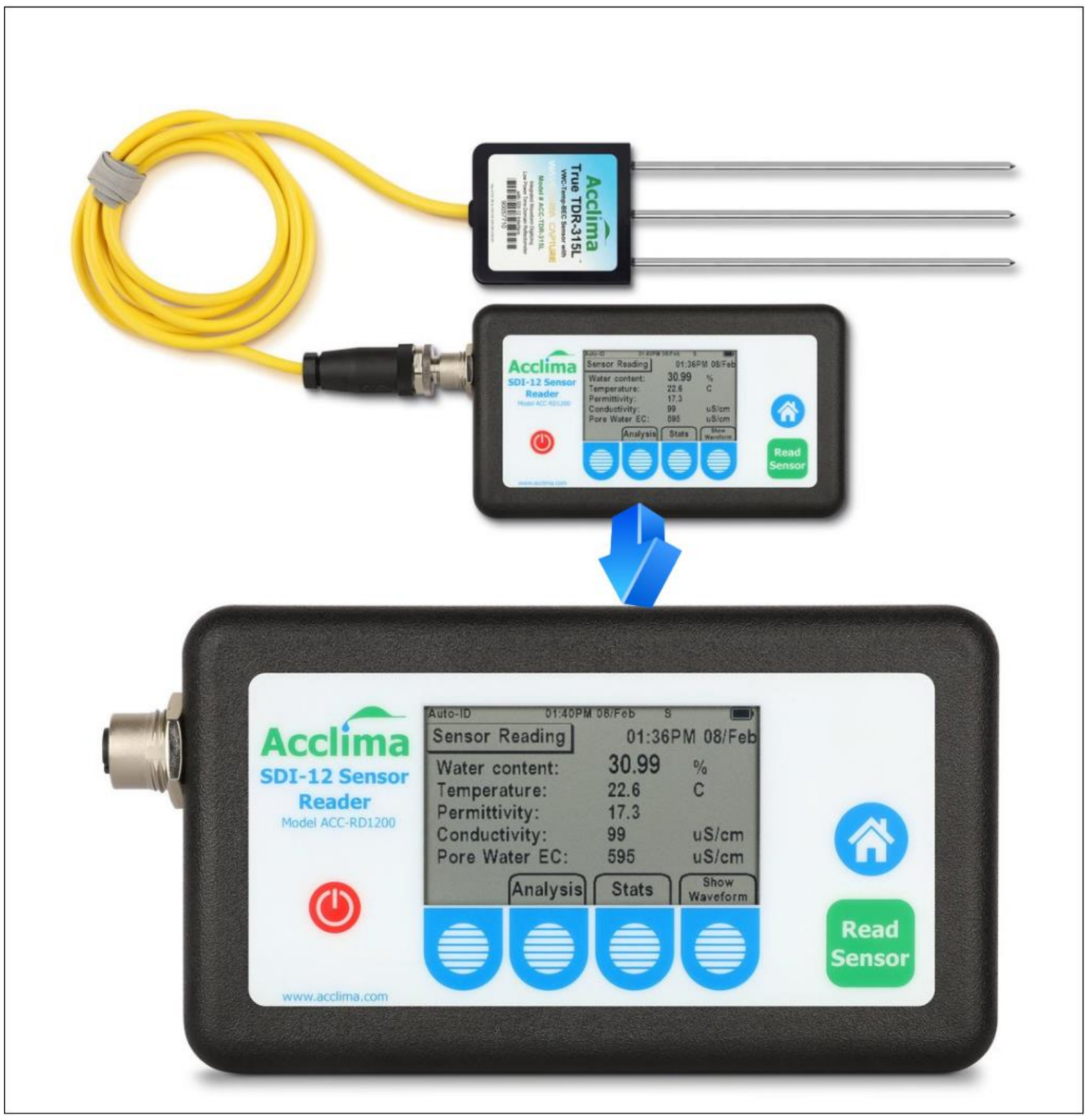

**Figure 2.** SDI-12 sensor reader.

**Table 4.** Statistics for actual moistures of the thirty-eight soil samples.

| | Moisture |
|---|---|
| Minimum | 0.71 |
| Maximum | 30.11 |
| Mean | 10.50 |
| Standard Deviation | 8.30 |

Two different iPhones were used to capture soil images. The first one was an iPhone 11 Pro, and the second was an iPhone 6s. The bottom part of each soil sample was captured for the image. We kept a distance of 50 cm while taking images of the soil samples using iPhones. The resolution of the soil image was 2688 × 1242 pixels for the iPhone 11 Pro and 1080 × 1920 pixels for the iPhone 6s. Four image sets were taken from 38 soil samples according to the iPhone versions and direct and indirect sunlight conditions during the image capture. Additionally, multiple images but not more than five instances for each image set were grabbed. The pictures of soil samples were captured without the flash of the mobile camera to standardize the experiment. Captured soil images were saved in PNG format for both devices. After reviewing all the images of soil samples, four datasets (Table 5) were finalized for the research based on mobile devices as well as direct and indirect sunlight. Table 6 shows examples of four soil images from each dataset with different moisture contents. These images indicate that a darker color refers to higher moisture content.

**Table 5.** A total number of instances for four datasets.

| Dataset No. | Description | Total Instances |
|---|---|---|
| Dataset 01 | Images were taken with the iPhone 6s in direct sunlight | 171 |
| Dataset 02 | Images were taken with the iPhone 6s in indirect sunlight | 186 |
| Dataset 03 | Images were taken with the iPhone 11 Pro in direct sunlight | 135 |
| Dataset 04 | Images were taken with the iPhone 11 Pro in indirect sunlight | 137 |

White balance is used in a camera to adjust the image colors with the light source color so that white objects appear in neutral white. Usually, a specific light source is not included in the photographic system, but a light source is vital to avoid color casts during the image capture [10,26]. Subjects can be illuminated with various light sources, such as sunlight, incandescent bulbs, and fluorescent lighting. The proper white balance prevents color distortion, and the color of the illumination source is essential for applying the white balance correctly [7]. Generally, a digital camera has a presetting option for adjusting the white balance [7,8]. For example, the authors in [7,10] utilized the gray card for adjusting the white balance setting in a Nikon digital camera with fluorescent lighting before taking soil sample images. However, iPhone cameras significantly differ from digital cameras in image processing. A digital camera captures a raw image without any modification. In contrast, iPhone images undergo various automated post-processing adjustments, including color correction, white balance, color interpolation, gamma correction, compression, and so on [27]. A digital camera's raw image holds significantly more metadata than a smartphone camera, which is helpful for manual post-processing [26]. In an iPhone, the images are recorded in metadata (original images) as well as enhanced images through post-processing [26]. Since an iPhone automatically deals with numerous post-processing adjustments, it changes the actual object posture for a better viewing experience. In this case, the iPhone has no white balance issue because the automatic white balance adjustment is made in an iPhone device when capturing an image [28].

**Table 6.** Four soil images of each dataset with different moisture contents.

| | Dataset 01 | Dataset 02 | Dataset 03 | Dataset 04 |
|---|---|---|---|---|
| Sample 9 (0.71% Moisture) | 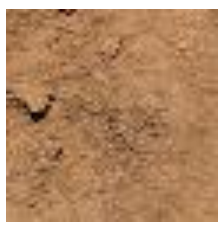 | 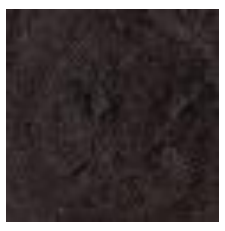 | 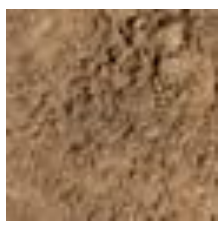 | 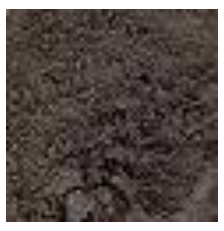 |
| Sample 15 (14.54% Moisture) | 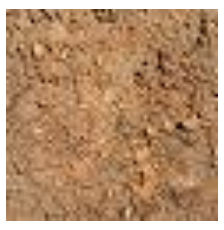 | 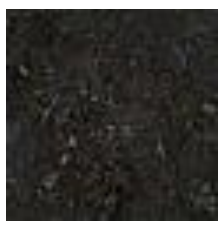 | 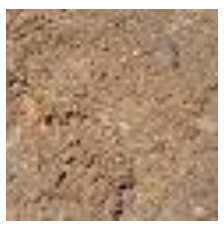 | 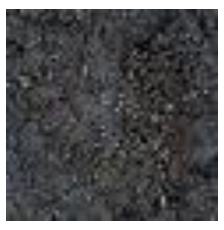 |
| Sample 33 (25.04% Moisture) | 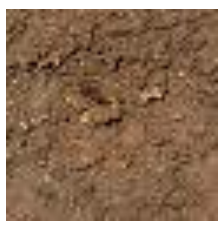 | 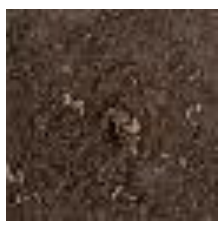 | 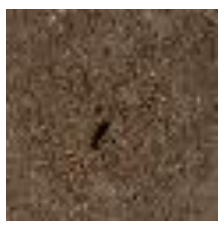 | 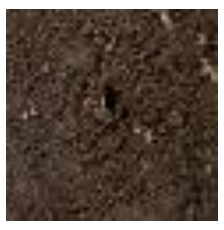 |
| Sample 13 (30.11% Moisture) | 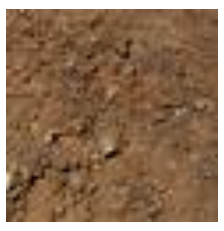 | 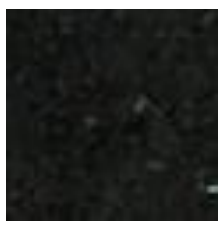 | 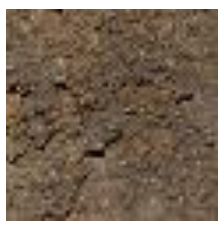 | 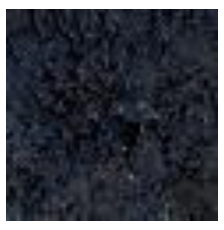 |

Researchers used various techniques to identify and remove multiple non-soil parts such as gravel, root, shading, and reflection of water or light. The authors in [11] selected a pixel intensity value for the sample images to identify non-soil parts that could not meet the value. In [5,6,8], the authors set a range of lowest and highest pixels to identify and remove the outlier pixels. In this study, non-soil particles such as gravel and root were identified and removed from the soil while capturing an image, as they can hamper the image quality.

A Light Meter (LM) is commonly used to measure the amount of light falling on a surface during image capture. The measurement of light intensity is essential for understanding whether a particular light source provides enough light for an intended surface. This meter works on an image cell to forgather light and convert the light into electricity, allowing the Lux value to be computed [29]. Light intensity is measured in Fc (lumens per square foot) or Lux (lumens per square meter). Even though multiple handheld devices are available in the market for the measurement of light intensity, an app named 'Light Meter' was operated using an iPhone's back camera to document the light intensity levels during the image capture in this study. During the collection of Fc and Lux values using the 'Light Meter' app, we kept the iPhone's camera vertical at 10 cm above the soil sample. Although Fc and Lux were collected along with the soil images, only Lux was implemented as an additional input parameter in the machine learning models.

### *3.2. Soil Image Analysis*

Although an effective image acquisition verifies the image quality, the correct image analysis methods draw crucial image information and are essential to computer vision applications [11]. Choosing a suitable image analysis technique can ensure the extraction of vital information from the images. Several researchers used ImageJ software written in Java to edit, analyze, and crop an image [5]. Similarly, the authors in [7-9] and [11] utilized MATLAB software for image processing. In this study, we carefully captured soil images using smartphones in direct and indirect sunlight during the fieldwork. However, several images were taken incorrectly (i.e., blurred)—these were manually identified and discarded. Then, the rest of the soil images were cropped squarely to remove their background. Finally, the cropped images were further cropped to 120 × 120 pixels using the 'image' class of the 'Pillow' library in Python.

Color spaces are utilized to define the range of colors. RGB, HSV, and monochrome color spaces are conventional to extract the values of the images. Depending on the amount of water in a soil sample, different colors are displayed by the reflection of electromagnetic energy in the soil [5,7,11]. In this case, the relationship between the color space values and the soil moisture manifested that the soils grew darker as moisture increased. The authors in [22] extracted only the RGB color space to calculate the median. To perceive the mean, the authors in [6,8] used the RGB color space. In another color space, HSV values were calculated by the authors in [5,9,10] with RGB. The authors in [11] took advantage of RGB, HSV, and monochrome color spaces to extract the mean and median values of the soil images. In [7], the authors applied RGB, HSV, and panchromatic color spaces to get the median values. Since the authors in [5,9] found that RGB had higher prediction accuracy than other color spaces in their research, only the RGB color space was selected to assess regression models for this study. In this case, only the mean values of the color space were used because it includes all the 120 × 120 pixels in the calculation. After the image segmentation, the values of the RGB color space were run to compute the mean values.

The correlation between each parameter (RGB color space values and Lux values) and actual soil moisture is illustrated in Figures 3–6. These figures indicate that R, G, and B variables negatively correlate with the moisture variable. On the other hand, the relationship between Lux values and the soil moisture presents a negative correlation for datasets 01 and 03 but a positive correlation for datasets 02 and 04. It means that a positive correlation occurs between the Lux value and the moisture value when the images are taken in direct sunlight, but a reverse reflection for the indirect sunlight images. However, the proportion of variances ($R^2$ error) is close to 1, which leads to fitting the regression line.

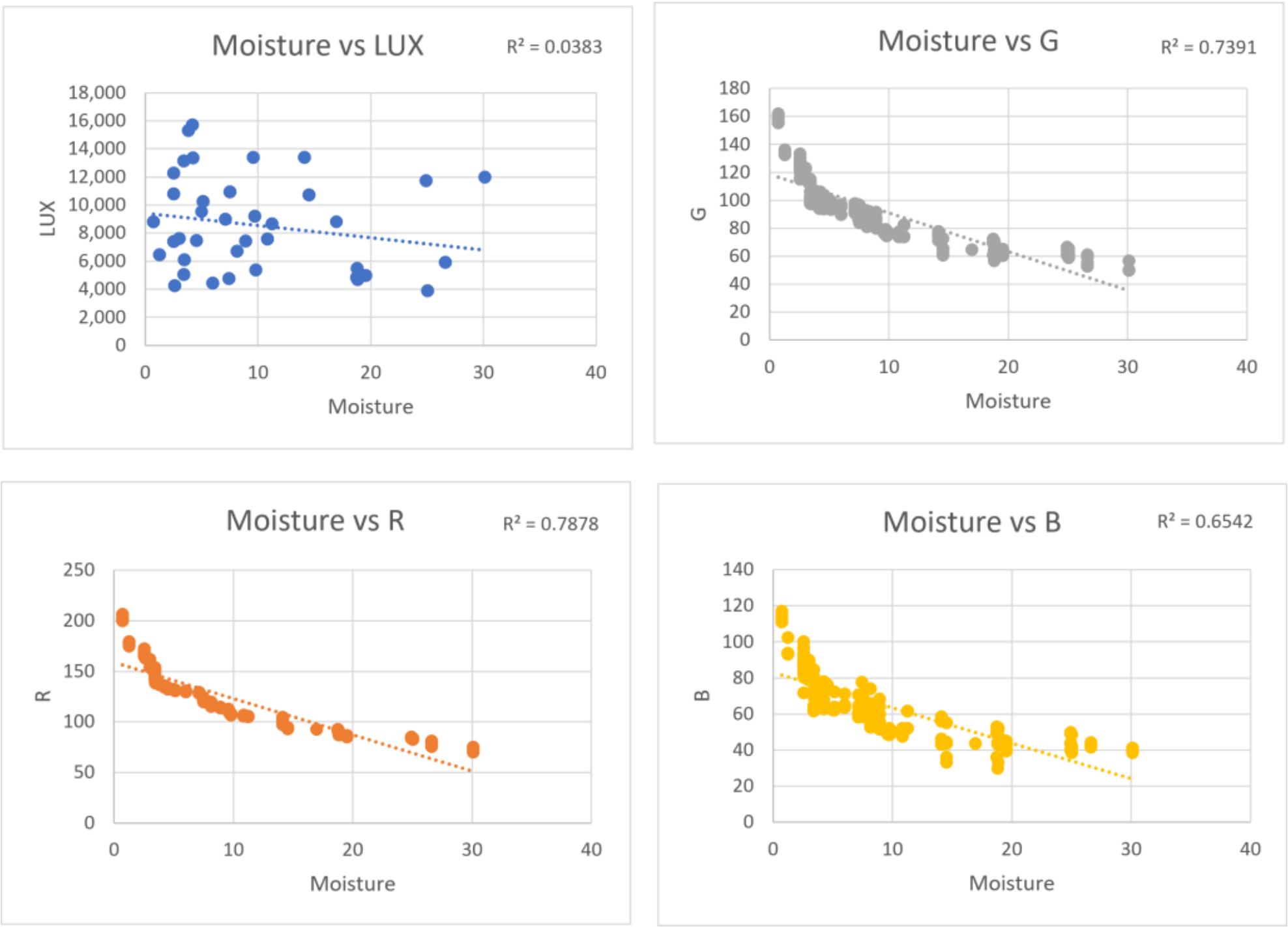

**Figure 3.** Correlations of RGB and Lux values with soil moisture for dataset 01.

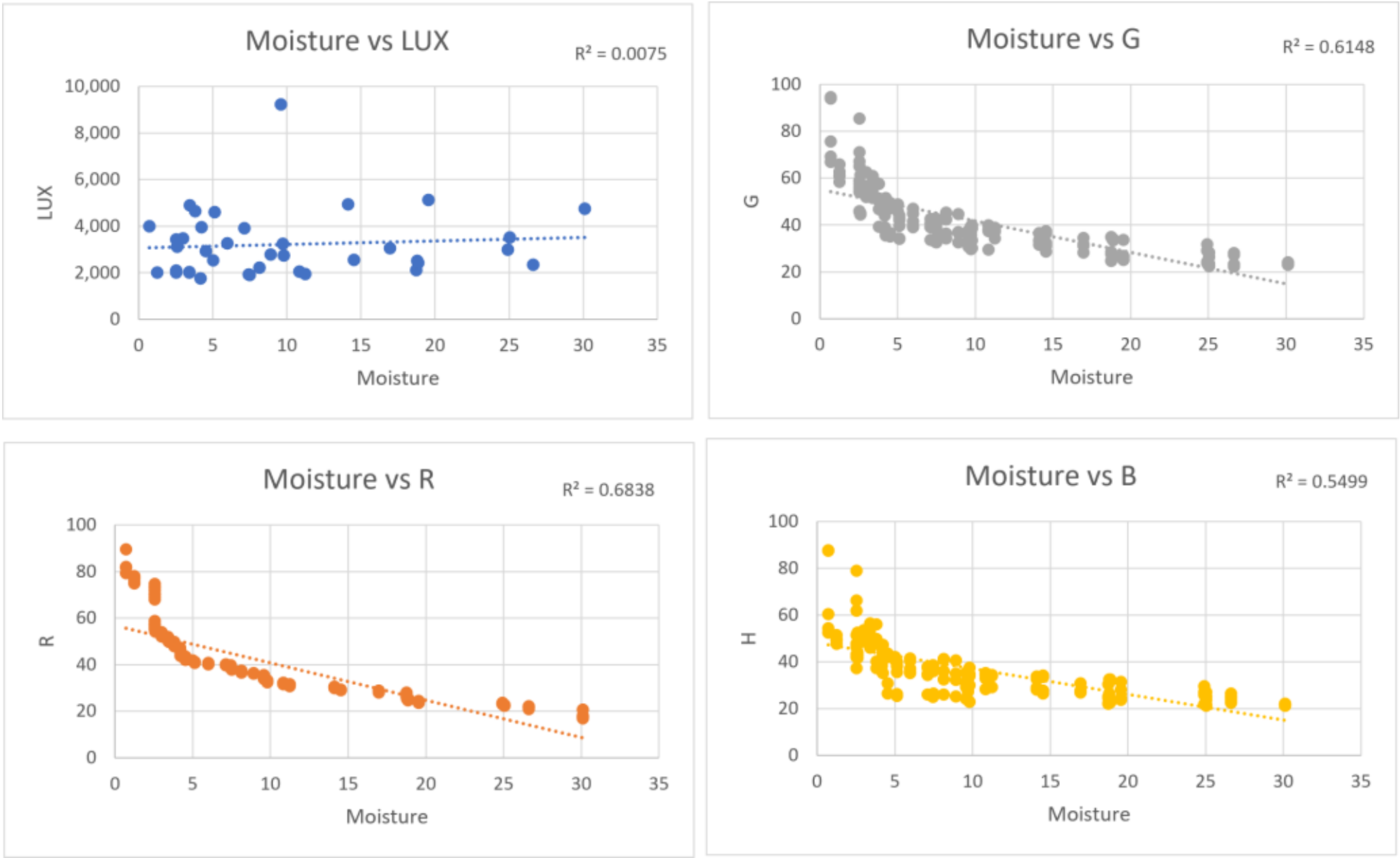

**Figure 4.** Correlations of RGB and Lux values with soil moisture for dataset 02.

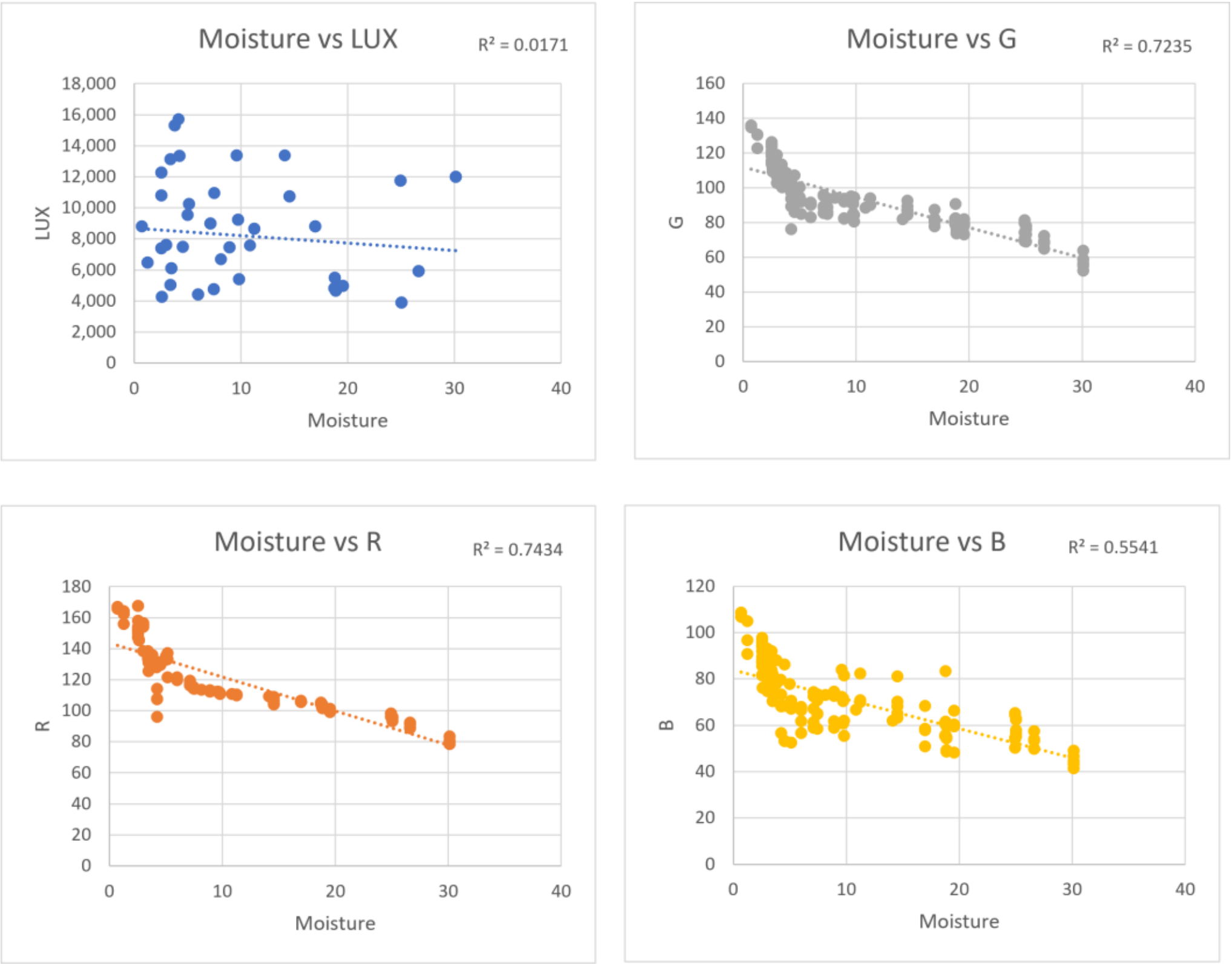

**Figure 5.** Correlations of RGB and Lux values with soil moisture for dataset 03.

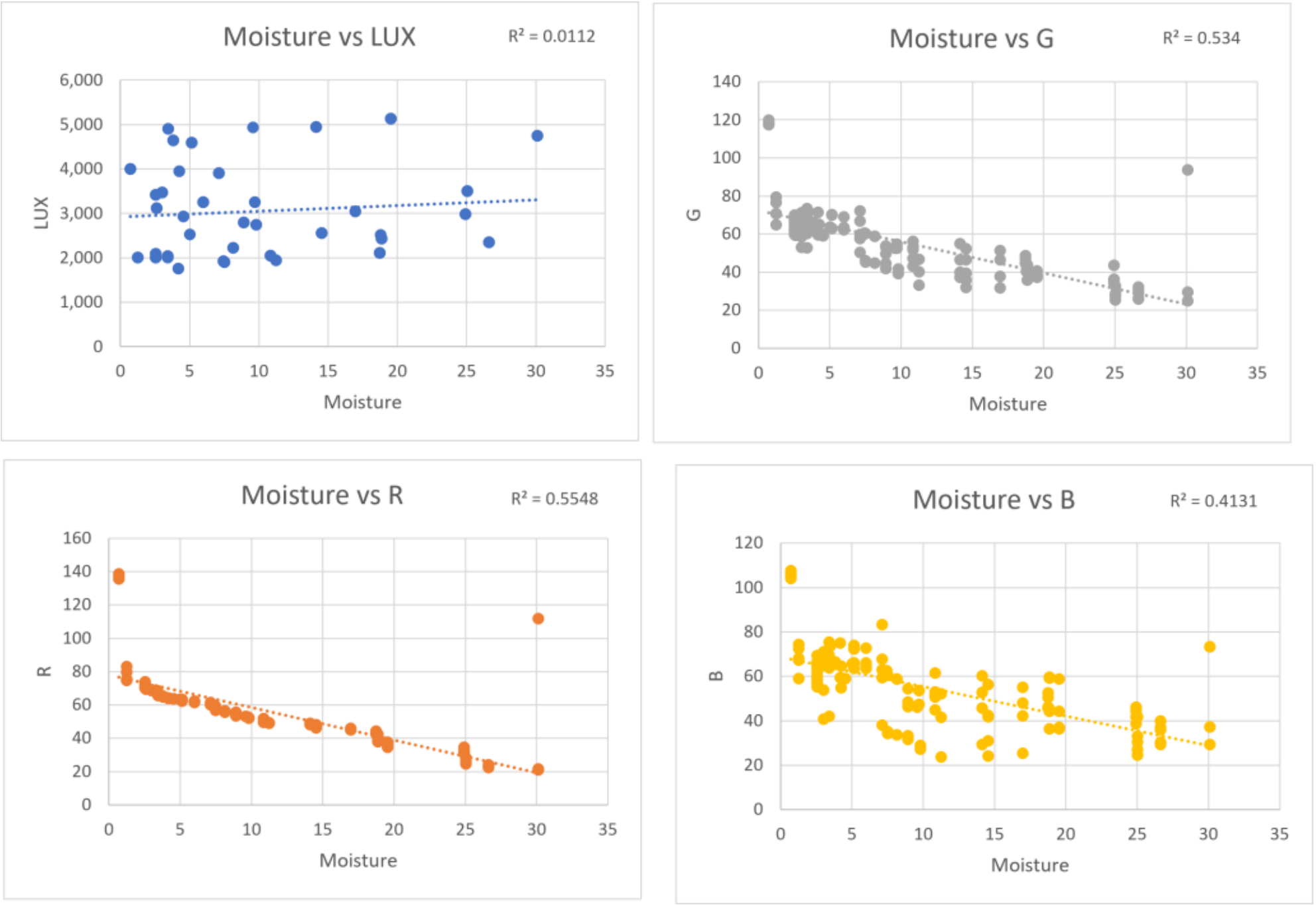

**Figure 6.** Correlations of RGB and Lux values with soil moisture for dataset 04.

### *3.3. Machine Learning Models*

ML is the scientific study of computational algorithms, which have constructed a model based on sample data, known as training data used for forecasting [30]. Therefore, ML allows a machine to learn with no explicit programming. ML studies present a variety of challenges when it comes to constructing high-performance regression models.

Therefore, it is crucial to select the appropriate ML algorithms for regression and the volume of data that needs to be handled by the algorithms. In the research, the ML process was split into four stages: data collection, known as row dataset; data cleaning with feature engineering; model building; and model evaluation, as illustrated in Figure 7. Several ML models were implemented in this research, including MLR, nonlinear Support Vector Regression (SVR), and CNN for understanding the prediction of soil moisture.

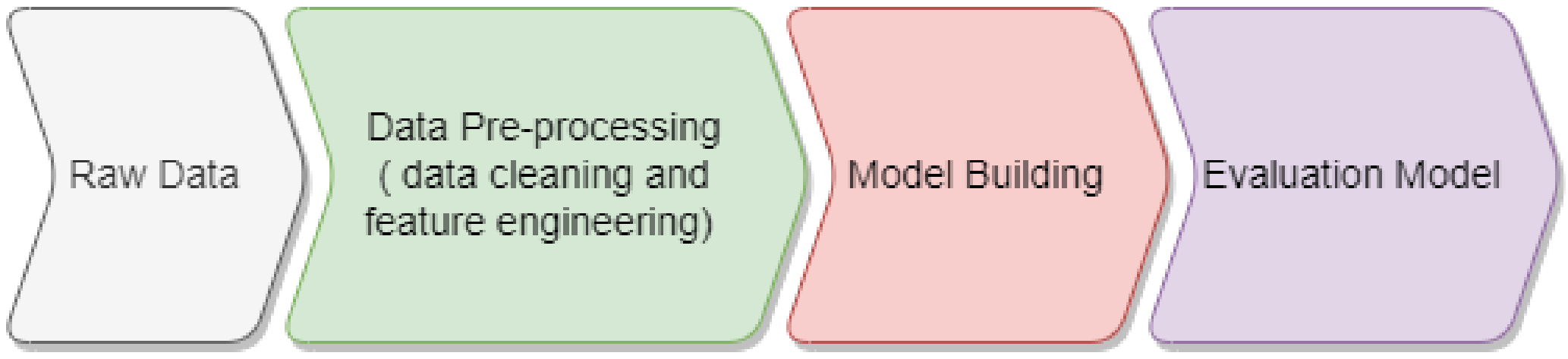

**Figure 7.** A machine learning process.

MLR is a popular ML model for prediction. It evaluates the relationship between one dependent variable and more than one independent variable. Multiple linear regression determines whether the datasets can meet certain assumptions. These assumptions are lost data analysis, multivariate normality, multivariate linearity, freedom from extreme values, and ties between independent variables. At first, these are analyzed one at a time. Then the regression analysis is evaluated with the data that satisfies those assumptions. In the MLR, more than one independent variable is assumed to have a linear relationship with one dependent variable, which attempts to reduce residual error by adjusting all data points in a straight line [31]. The below formula is used for multiple linear regression.

$$y = \beta_0 + \beta_1 x_1 + \cdots + \beta_n x_n + \varepsilon \quad (1)$$

where $y$ is a dependent variable, $x_i$ is an independent variable, $\beta_i$ is a parameter, and $\varepsilon$ is an error.

SVM regression, also referred to as Support Vector Regression (SVR), is adapted as a prediction tool. In this case, a hyperplane that lies close to as many points as possible is constructed [21,31]. A kernel is used as a parameter to determine a hyperplane in the higher dimensional space [32]. An enhanced dimension is necessary when it is challenging to discover a separating hyperplane in a particular dimension. Nonlinear mapping is utilized by SVM to map input vectors into high-dimensional feature spaces. In nonlinear regression, data are fitted to a model and subsequently expressed as a mathematical function.

Deep Learning (DL) mimics human brain decision-making and has been successfully applied for regression [33]. A neural network of over three layers can be considered a deep learning algorithm. One of the deep learning models is a CNN, which is proven for image processing tasks. It consists of an input layer, convolution layer, pooling layer, and fully connected layer. The convolution layer contains multiple filters known as kernels. The convolution operator has parameters such as filter size, padding, stride, dilation, and activation function. The filter scans the whole image, and an activation function is applied to the output to clarify any nonlinearity. Several deep neural networks are feed-forward that have an input to the output direction flow. However, a deep neuron network can be trained by backpropagation, which moves in the opposite direction from the output to the input.

In previous research, several ML models were utilized for soil moisture estimation. For example, the LR [7,10,11], MLR [5,7,9], SVR [8,11], ANN [5,6,8], PLS [6], Decision/Regression Tree [11], GPR [11], Random Forest [11], and Cubist [11] models were implemented singly or jointly for estimating soil moisture from soil images. Often, soil images were converted into color space(s) utilizing input parameters. In this regard, the authors in [11] extracted 22 features of color and texture from RGB, HSV, and monochrome images of the soil samples to use as input variables. Similarly, the authors in [5,9,10] utilized the RGB and HSV as input variables for training the models in their research. In [7], the authors used RGB, HSV, and DN (Digital Number) values as input parameters. However, the authors in [8] applied only RGB values as inputs to train their models. On the other hand, the authors in [6] used nine input variables consisting of mean RGB value and site-specific data, including land cover, vegetative cover, canopy cover, altitude, profile depth, slope, landform, and topography for their research.

MLR, nonlinear SVR, and CNN were implemented individually or combined with other models in previous research. Still, comparing these three models has not yet been done for soil moisture prediction. Therefore, this study compared MLR, SVR, and CNN with minimum input parameters or variables applied to the models to avoid complexity. In this study, the mean values of R, G, and B from the RGB color space were used as the independent or input variables for MLR and SVR, where soil moisture percentage was a dependent variable or outcome. In the case of CNN, the soil imaging was implemented directly instead of RGB color space as the primary input variable. Another additional input variable used for training the models was Lux.

### *3.4. Cross-Validation Techniques and Evaluation Metrics*

Cross-validation is a method of evaluating ML models. Most of the researchers used a holdout cross-validation technique to evaluate ML models during soil moisture prediction, followed by a k-fold cross-validation technique. In this research, we applied multiple cross-validation techniques such as holdout, k-fold, and leave-one-out to assess the performance of various ML models and understand their effects on soil moisture accuracy.

The holdout cross-validation technique divides data into multiple instances, such as training, validation, and testing sets. A training set is used to achieve the model parameter values; a validation set is used to measure the performance; a testing set is used for unbiased generalization performance estimation. In [5,8], the authors used the holdout cross-validation technique for model assessment. The authors in [8] divided the data vectors into the training subset (70%), validation subset (15%), and test subset (15%) for the MLP method and training subset (85%), and test subset (15%) for the SVR method. Similarly, the authors in [5] used 85% of the data for training and 15% for validation for evaluating the ANN model. This study used the holdout cross-validation technique, where 70% of instances were utilized for training purposes and the rest (30%) for testing.

A k-fold cross-validation technique is used for evaluating ML models over a limited sample of data. In this case, the value of the K parameter determines how the sample data are divided into several groups. Then the model is trained on K-1 subsets, and the assessment is done on the new subset [34]. The authors in [6,11] used 10-fold cross-validation in their research. This study used a k-fold cross-validation technique, and the data were divided into ten equal-sized parts. In this paper, the K value was 10.

We also applied a leave-one-out approach in this study to understand the models' effectiveness, where only a single observation is present for validation. In the leave-one-out practice, the whole training set is used once at a time and combines all results to estimate error [34,35].

Evaluation metrics were used to calculate the difference between actual and predicted values for understanding model performance. Many different evaluation metrics exist, but only some were used for soil moisture regression. Among them, the Coefficient of Determination ($R^2$) was common [5,6,7–11,22]. The authors in [5–7,10,11] adopted the Root Mean Square Error (RMSE) for evaluation. Similarly, Lin's Concordance Correlation Coefficient (LCCC) [11], mean of the residuals (bias) [11], Mean Squared Error (MSE) [8], Ratio of Performance to Deviation (RPD) [11], and Ratio of Performance to Interquartile Distance (RPIQ) [11], have also been used for soil moisture estimation. Mean Absolute Error (MAE), RMSE, and $R^2$ were used to predict the moisture of soil samples in this research because they are the conventional metrics to measure accuracy.

MAE is used to determine the sum of the absolute value of error. Firstly, it finds the total value from the difference between actual and predicted outcomes, eliminating the negative sign. After that, it calculates the mean value using calculated absolute values. Therefore, the values of MAE change linearly due to finding the absolute value from the difference between the actual and predicted results [36]. The below formula is implemented for MAE.

$$MAE = \frac{1}{n}\sum_{i=1}^{n} |y_i - x_i| \quad (2)$$

Where $n$ corresponds to the number of samples within the test dataset, $y_i$ is the prediction, and $x_i$ is the true value.

RMSE is implemented to compare a predicted value with an observed value. RMSE first finds the squared difference between the actual value and predicted value, then calculates the mean value of the squared difference. Finally, it performs a root operation on the mean value. Finding the squared difference between the actual and predicted results eliminates the negative value. As a result, a positive error value is generated from the model performance. RMSE can handle the larger error values by magnifying the mean score because of the square error values [36]. For this reason, RMSE is invoked to identify the more significant error rates. The below formula is used for RMSE.

$$RMSE = \sqrt{\frac{\sum_{n=1}^{N}(\hat{r_n} - r_n)^2}{N}} \quad (3)$$

where $\hat{r_n}$ indicates the rating of prediction, $r_n$ denotes the actual rating in a testing dataset, and $N$ refers to the sample numbers in the testing dataset.

The $R^2$ error is related to the variance of actual and predicted values based on samples, and the error range is between $-\infty$ and 1 in regression analysis [37,38]. The $R^2$ of 1 means that the movements of independent variable(s) entirely explain the movements of a dependent variable. On the other hand, a zero value of $R^2$ indicates that a model is not adapted by samples [31]. The $R^2$ value also can be negative when the selected model does not follow the data. In general, a value of $R^2$ close to 1 is satisfactory for a model [37]. It is easy to estimate the differences between each number from the mean as the $R^2$ error metric uses variance to determine the result. The below formula is used for $R^2$.

$$R^2 = 1 - (\frac{RSS}{TSS}) \quad (4)$$

$$RSS = \sum_{i=1}^{n}(y_i - \hat{y_i})^2 \quad (5)$$

$$TSS = \sum_{i=1}^{n}(y_i - \bar{y})^2 \quad (6)$$

where TSS is the sum of squares together; RSS indicates the residual sum for squares; $y_i$ is the actual value; $\hat{y_i}$ is the predicted value; and $\bar{y}$ is the mean of the actual values.

### *3.5. Experimental Setting*

#### 3.5.1. Models Setting

In this research, three ML models (the MLR model, the SVR model, and CNN) were implemented to evaluate their performance in estimating soil moisture. The first two ML models were tested against the RGB color space (i.e., mean values of R, G, and B) of the soil images, but the CNN model was reviewed based on the soil imagery. In both cases, Lux was used as an additional input parameter.

During the system development, a Python library known as scikit-learn was implemented for the MLR and SVR models to train the datasets. However, this research exploited the Python library called TensorFlow for the CNN model.

Linear regression is among the well-known algorithms in statistics and ML. A multiple linear regression model was implemented to predict soil moisture using the following expression.

$$y = a + (R_{mean} \times a1) + (G_{mean} \times a2) + (B_{mean} \times a3) + (Lux \times a4) \quad (7)$$

where $y$ represents the dependent variable; $R_{mean}$, $G_{mean}$, and $B_{mean}$ are the first three independent variables that are the mean values of the RGB color space; $Lux$ is another independent variable; $a$ is the intercept; and $a1$, $a2$, $a3$, and $a4$ are biases for each input, and at the beginning, values of the individual coefficient start from random initialization.

An SVR model utilized regression analysis to predict moisture levels in this research. The mean of the RGB values of the soil images, and the values of Lux, were entered as inputs in the model. Although SVR is a distance-dependent model, normalized inputs were offered to this model to predict the performance of scaling data. Since the datasets in this research included nonlinear data, whose trend was curved, the SVR model utilized its kernel trick, which helped to transform the nonlinear data into a high-dimensional feature space where each input represents a point in the space. Based on the high-dimensional space, the hyperplane found the optimal boundary to predict the possible output.

The DL model, specifically the CNN, was employed for soil moisture assessment for this study. In this regard, the TensorFlow package in the Jupyter Notebook web tool was installed to build a prediction model to estimate soil moisture from soil images and the value of the solar light status during image capture. The input parameters and precision of several deep neural networks were reviewed to construct an efficient CNN model. The standard seven-layer feed-forward network was used; it had two input layers. One input layer accepted the pixels of soil images directly, and the other accepted Lux as an additional parameter. Based on the inputs, soil moisture was predicted in the output. The CNN architecture of this research is illustrated in Figure 8.

The DL model implemented the three convolutional layers among the seven layers to extract the feature from each input image. Each was followed by one max pooling layer and one batch normalization layer to extract the features from the images. In addition, the global average pooling layer was added at the last convolutional layer, and three dense layers were utilized to introduce the additional parameter (Lux value) to the neural network model. It is noted that 32, 64, and 128 filters were used in three consecutive convolutional layers. In the CNN model, the kernel size was (3,3), the padding method was 'same,' and the stride was (1,1). Here, 'filters' were used to create channels that learn specific pixels while training the model. These filter attributes accepted only integer values, which were numerous channels in one convolutional layer. Kernel size indicated the filter channel size precisely as 2d metrics of each filter. Padding and stride helped the convolutional layer to select the specific pixel. 'ReLU' was chosen for the activation function in every hidden layer because it eliminated negatively weighted neurons from the model as it follows the formula $z = Max(0, z)$, where $z$ is the neuron weight. The 'ReLU' activation function is useful for adding a regularization (dropout) layer.

In the output layer, the 'Tanh' activation function was used to generate values in the range of [−1, 1]. The number of parameters was 104,515. Among them, 104,323 parameters were trainable, and 192 parameters were non-trainable. To spot the global minima, the 'Adam' optimizer function was utilized with a learning rate of 0.001.

We used 629 soil images in this research, and the total dataset was split into train and validation. To avoid overfitting and underfitting issues, the CNN model was trained multiple times by varying the number of layers and neurons in a layer. Regularization prevents the model from overfitting the training data [23]. For this reason, we added a 0.3 value with the regularization or dropout layer to block the neurons. Moreover, three additional parameters were introduced while compiling the model. These are checkpoints to save the best model, early stopping to control the model before being overfitted, and reducing the learning rate to help the model to converge correctly.

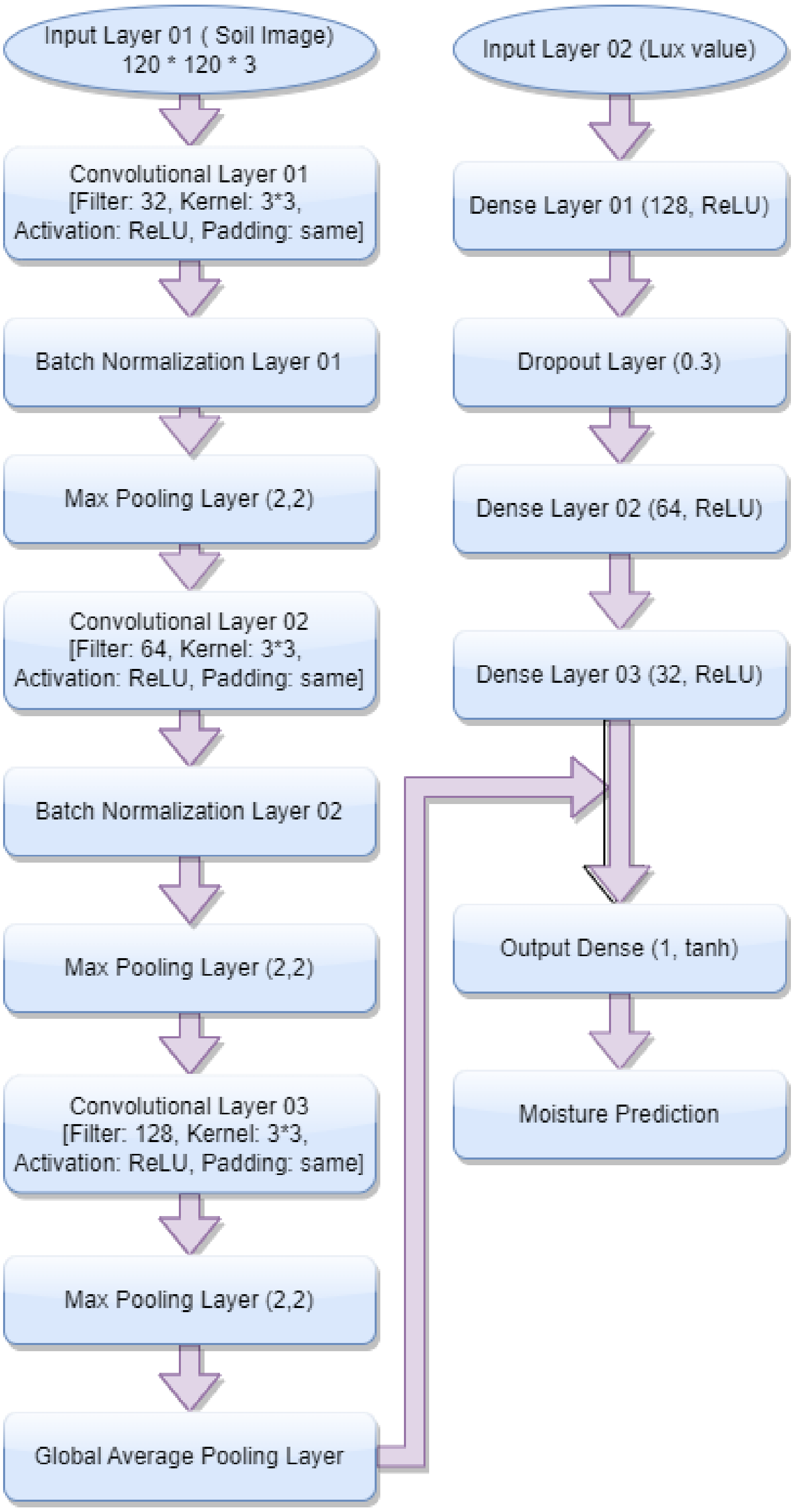

**Figure 8.** Convolutional neural network architecture.

### 3.5.2. Feature Scaling

After checking the linearity of collected datasets, a few scaling techniques were performed to scale the datasets. Primarily, ambiguity and the presence of negative values in the datasets were focused on and scaled using the appropriate scaling methodologies. In this regard, the 'MinMax' scaler, Standard scaler, and 'MaxAbs' scaler techniques were used on the datasets. A Python library known as scikit-learn was implemented for the scaling technologies in this research. We used lux as an additional feature whose values were between 1761 and 9237 in indirect sunlight and between 15,717 and 3893 in direct sunlight.

We implemented the 'MinMax' scaler to normalize the lux values between 0 and 1. This scaling method properly fits for datasets in this research with no negative values or ambiguity. The following formula was applied during the 'MinMax' scaling.

$$z = (x - \min(x))/(\max(x) - \min(x)) \quad (8)$$

where z is a scaled value, $min$ is a minimum value, and $max$ is the maximum value of an x attribute in the dataset.

The standard scaling technique was implemented to normalize the datasets where the mean of observed values remains zero, and the standard deviation persists at one. However, the standard scaling method generated some negative values in the datasets, which were not expected. The following formula was utilized during standard scaling.

$$z = (x - m)/s \quad (9)$$

where $z$ is the scaled value, $x$ is to be the scaled value, m is the mean value of the dataset, and $s$ is the standard deviation of the dataset values.

Lastly, 'MaxAbs' scaling was used in the datasets, generating a similar result to the 'MinMax' scaling method; however, after generating a graph between 'MinMax' and 'MaxAbs,' a lower trend is suspected in the 'MaxAbs' Scaler concerning 'MinMax' Scaler. As a result, the outcome of 'MinMax' scaling was considered instead of 'MaxAbs' scaling for this research. The following formula was implemented during 'MaxAbs' scaling.

$$y_i = x_i/absmax(x) \quad (10)$$

where $absmax(\ )$ is used to determine the maximum value in an attribute by neglecting the negative sign.

## 4. Results and Discussion

The MLR, SVR, and CNN models were used to predict soil moisture. As part of this research, three cross-validation techniques (holdout, k-fold, and leave-one-out) were implemented to understand the performance of the models concerning the four datasets. These datasets were based on the usage of iPhones under direct or indirect sunlight while capturing images from the soil samples. For each validation technique, three different error metrics (MAE, RMSE, and $R^2$) were computed. The comparison results of the ML models using different validation techniques on individual datasets are included in Tables 7–9. Moreover, the models were trained by one dataset and assessed by the other three datasets, as documented in Tables 10–13. Furthermore, the datasets were combined into one dataset, and then evaluation metrics applied for each validation technique are listed in Table 14. Finally, the comparison of sunlight conditions and the various smartphone devices for the higher accuracy model of this research are represented in Figures 9 and 10, respectively.

**Table 7.** Accuracy assessment using holdout cross-validation (the ratio of the training and testing dataset is 70:30).

| Dataset | Model | MAE | RMSE | $R^2$ |
|---|---|---|---|---|
| Dataset 01 | MLR | 0.45 | 0.26 | 0.09 |
| | SVR | 0.49 | 0.31 | 0.01 |
| | CNN | 0.50 | 0.29 | −0.52 |
| Dataset 02 | MLR | 0.35 | 0.15 | 0.60 |
| | SVR | 0.54 | 0.37 | −0.43 |
| | CNN | 0.58 | 0.43 | −1.38 |
| Dataset 03 | MLR | 0.45 | 0.26 | 0.06 |
| | SVR | 0.51 | 0.33 | −0.13 |
| | CNN | 0.57 | 0.42 | −1.40 |
| | MLR | 0.39 | 0.18 | 0.54 |
| Dataset 04 | SVR | 0.47 | 0.30 | −0.09 |
| | CNN | 0.44 | 0.27 | −0.12 |

According to Table 7, the prediction error is minimum in MLR when soil images were taken in indirect sunlight using any of the iPhones (datasets 02 and 04), where MAE, RMSE, and $R^2$ are 0.35, 0.15, and 0.60, respectively. Correspondingly in the iPhone 6s, while MAE, RMSE, and $R^2$ are 0.39, 0.18, and 0.54, respectively, in the iPhone 11 Pro. Similarly, MLR performs better than other direct sunlight models (datasets 01 and 03). To exemplify, by splitting training and testing data to 70:30, the MLR model works better compared to the other models.

The values of the error matrices are listed in Table 8 using a 10-fold cross-validation method. In that case, the SVR model provides minimal error for both iPhones in direct and indirect sunlight. The preferable values of MAE, RMSE and $R^2$ are 0.05, 0.06, and 0.96, respectively, in indirect sunlight on the iPhone 6s for the SVR model.

**Table 8.** Accuracy assessment using k-fold cross-validation (here, the K value is 10).

| **Dataset** | **Model** | **MAE** | **RMSE** | $R^2$ |
|---|---|---|---|---|
| Dataset 01 | MLR | 0.21 | 0.26 | −0.12 |
| | SVR | 0.17 | 0.22 | 0.14 |
| | CNN | 0.56 | 0.39 | −3.61 |
| Dataset 02 | MLR | 0.13 | 0.16 | 0.65 |
| | SVR | 0.05 | 0.06 | 0.96 |
| | CNN | 0.47 | 0.27 | −3.96 |
| Dataset 03 | MLR | 0.21 | 0.26 | −0.17 |
| | SVR | 0.16 | 0.24 | 0.34 |
| | CNN | 0.50 | 0.30 | −4.53 |
| Dataset 04 | MLR | 0.14 | 0.19 | 0.44 |
| | SVR | 0.08 | 0.11 | 0.85 |
| | CNN | 0.50 | 0.30 | −4.53 |

**Table 9.** Accuracy assessment using leave-one-out cross-validation.

| **Dataset** | **Model** | **MAE** | **RMSE** | $R^2$ |
|---|---|---|---|---|
| Dataset 01 | MLR | 0.46 | 0.26 | 0.05 |
| | SVR | 0.40 | 0.22 | 0.31 |
| | CNN | 0.49 | 0.28 | −0.65 |
| Dataset 02 | MLR | 0.36 | 0.16 | 0.67 |
| | SVR | 0.22 | 0.06 | 0.95 |
| | CNN | 0.49 | 0.30 | −1.08 |
| Dataset 03 | MLR | 0.45 | 0.26 | 0.21 |
| | SVR | 0.40 | 0.24 | 0.34 |
| | CNN | 0.49 | 0.29 | −0.63 |
| Dataset 04 | MLR | 0.38 | 0.19 | 0.53 |
| | SVR | 0.27 | 0.10 | 0.88 |
| | CNN | 0.44 | 0.22 | −0.38 |

Table 9 lists the accuracy assessment using the leave-one-out cross-validation technique. The SVR model is better in this validation technique. In this case, the samples that were captured using the iPhone 6s under indirect sunlight exhibit a better outcome with MAE, RMSE, and $R^2$ values of 0.22, 0.06, and 0.95, respectively.

To evaluate the accuracy of soil moisture prediction between datasets, we tested a single dataset against the other datasets in this study. Table 10 lists the values of error matrices using dataset 01 versus other datasets. The SVR model exhibits a minimal error of MAE, RMSE, and $R^2$ of 0.45, 0.26, and 0.18, respectively.

**Table 10.** Accuracy assessment when trained with dataset 01 and tested with other datasets.

| Model | MAE | RMSE | $R^2$ |
|---|---|---|---|
| MLR | 0.48 | 0.28 | 0.02 |
| SVR | 0.45 | 0.26 | 0.18 |
| CNN | 0.48 | 0.28 | 0.04 |

In Table 11, this study listed the evaluation of the accuracy of dataset 02 against other datasets. The SVR performs better than MLR and CNN. The MAE value is 0.48; the RMSE is 0.28, and the $R^2$ is −0.03 for SVR.

**Table 11.** Accuracy assessment when trained with dataset 02 and tested with other datasets.

| Model | MAE | RMSE | $R^2$ |
|---|---|---|---|
| MLR | 0.94 | 0.90 | −12.93 |
| SVR | 0.48 | 0.28 | −0.03 |
| CNN | 0.59 | 0.45 | −1.58 |

SVR displays the higher result followed by MLR and CNN for dataset 03 versus other datasets, as demonstrated in Table 12. The better results of MAE, RMSE, and $R^2$ are 0.47, 0.28, and −0.06, respectively, for the SVR model.

**Table 12.** Accuracy assessment when trained with dataset 03 and tested with other datasets.

| Model | MAE | RMSE | $R^2$ |
|---|---|---|---|
| MLR | 0.50 | 0.29 | −0.09 |
| SVR | 0.47 | 0.28 | −0.06 |
| CNN | 0.53 | 0.32 | −0.37 |

Table 13 lists the accuracy assessment using dataset 04 against other datasets. For dataset 04, MLR and SVR present almost comparable results. In this case, the MAE and RMSE values are 0.48 and 0.32 for MLR and 0.49 and 0.32 for SVR. Similarly, the $R^2$ values are −0.38 and −0.37 for MLR and SVR, respectively.

**Table 13.** Accuracy assessment when trained with dataset 03 and tested with other datasets.

| Model | MAE | RMSE | $R^2$ |
|---|---|---|---|
| MLR | 0.48 | 0.32 | −0.38 |
| SVR | 0. 49 | 0.32 | −0.37 |
| CNN | 0.50 | 0.35 | −0.58 |

We combined all the datasets to estimate the accuracy level of soil moisture. The results of the combined datasets for the holdout, k-fold, and leave-one-out cross-validation techniques are listed in Table 14. For holdout cross-validation, MLR and SVR both perform better with 0.46, 0.26, and 0.18, and 0.46, 0.27, and 0.17 for MAE, RMSE and $R^2$, respectively. On the other hand, SVR is the superior technique for both 10-fold cross-validation and leave-one-out cross-validation. In this case, SVR obtains higher results for MAE, RMSE, and $R^2$ with values of 0.15, 0.20, and 0.48, respectively, at 10-fold cross-validation. SVR also performs better for leave-one-out cross-validation with values of 0.38, 0.20, and 0.50 for MAE, RMSE, and $R^2$, respectively.

**Table 14.** Accuracy assessment using combined datasets.

| Validation Technique | Model | MAE | RMSE | $R^2$ |
|---|---|---|---|---|
| Holdout cross-validation | MLR | 0.46 | 0.26 | 0.18 |
| | SVR | 0.46 | 0.27 | 0.17 |
| | CNN | 0.48 | 0.28 | 0.07 |
| K-fold Cross-Validation | MLR | 0.27 | 0.26 | 0.12 |
| | SVR | 0.15 | 0.20 | 0.48 |
| | CNN | 0.51 | 0.31 | −0.44 |
| | MLR | 0.45 | 0.26 | 0.12 |
| Leave-one-out cross-validation | SVR | 0.38 | 0.20 | 0.50 |
| | CNN | 0.51 | 0.32 | −0.73 |

Although this study identified that the SVR model is a better performer using the research's datasets, there is a slight effect on the model's efficiency based on images captured in direct and indirect sunlight (Figure 9). However, the error is comparatively low when a model is trained using indirect sunlight images.

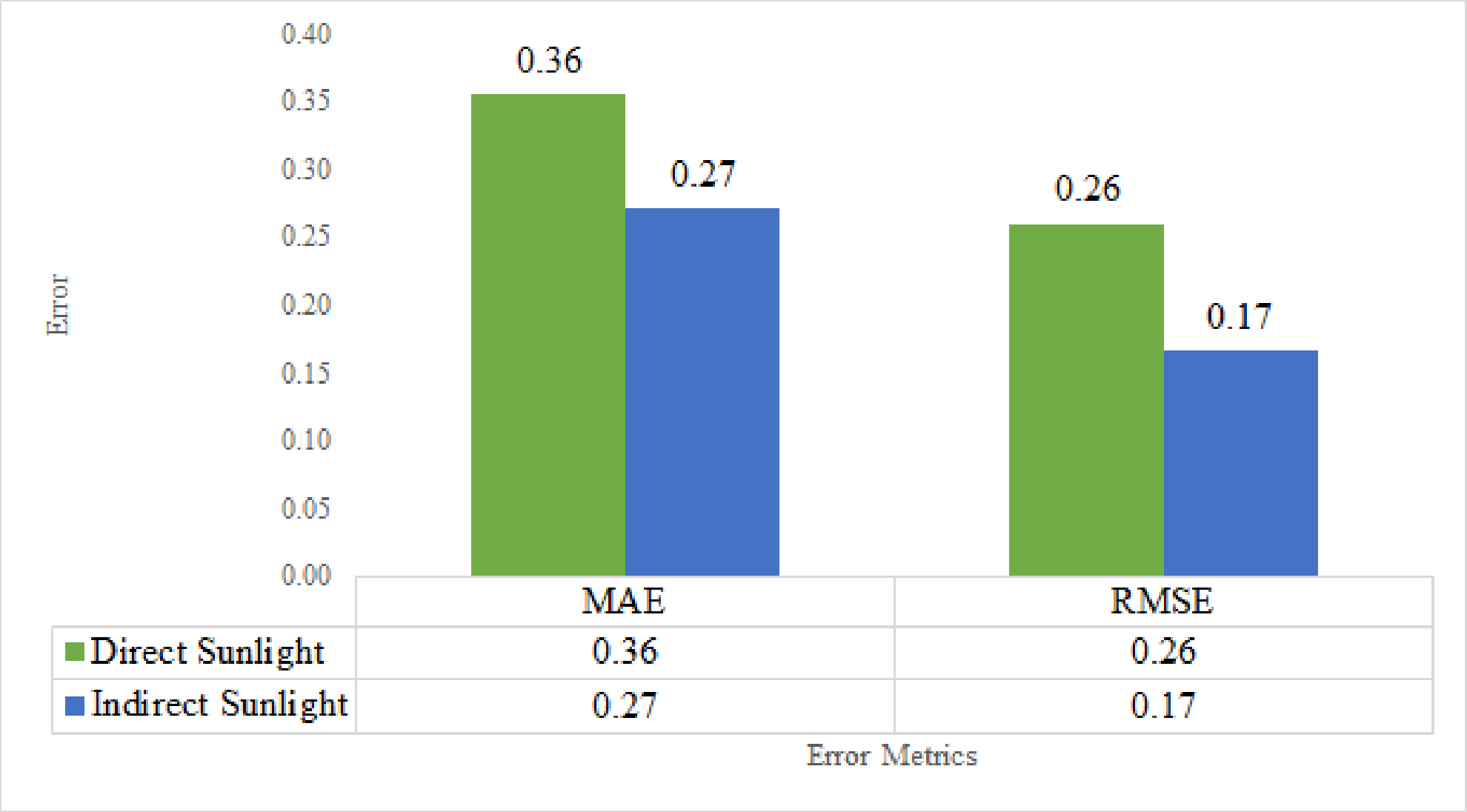

**Figure 9.** The error comparison between direct sunlight and indirect sunlight images in the SVR model.

The summary of the SVR model is drawn based on results from the iPhone 6s and iPhone 11 Pro in Figure 10. Here, the error metric depicts that the iPhone 6s exhibits better prediction in MAE and RMSE while running the SVR model because of more sample data for the iPhone 6s than the iPhone 11 Pro. However, based on the prediction results in this research, there are no significant distinctions between the two devices.

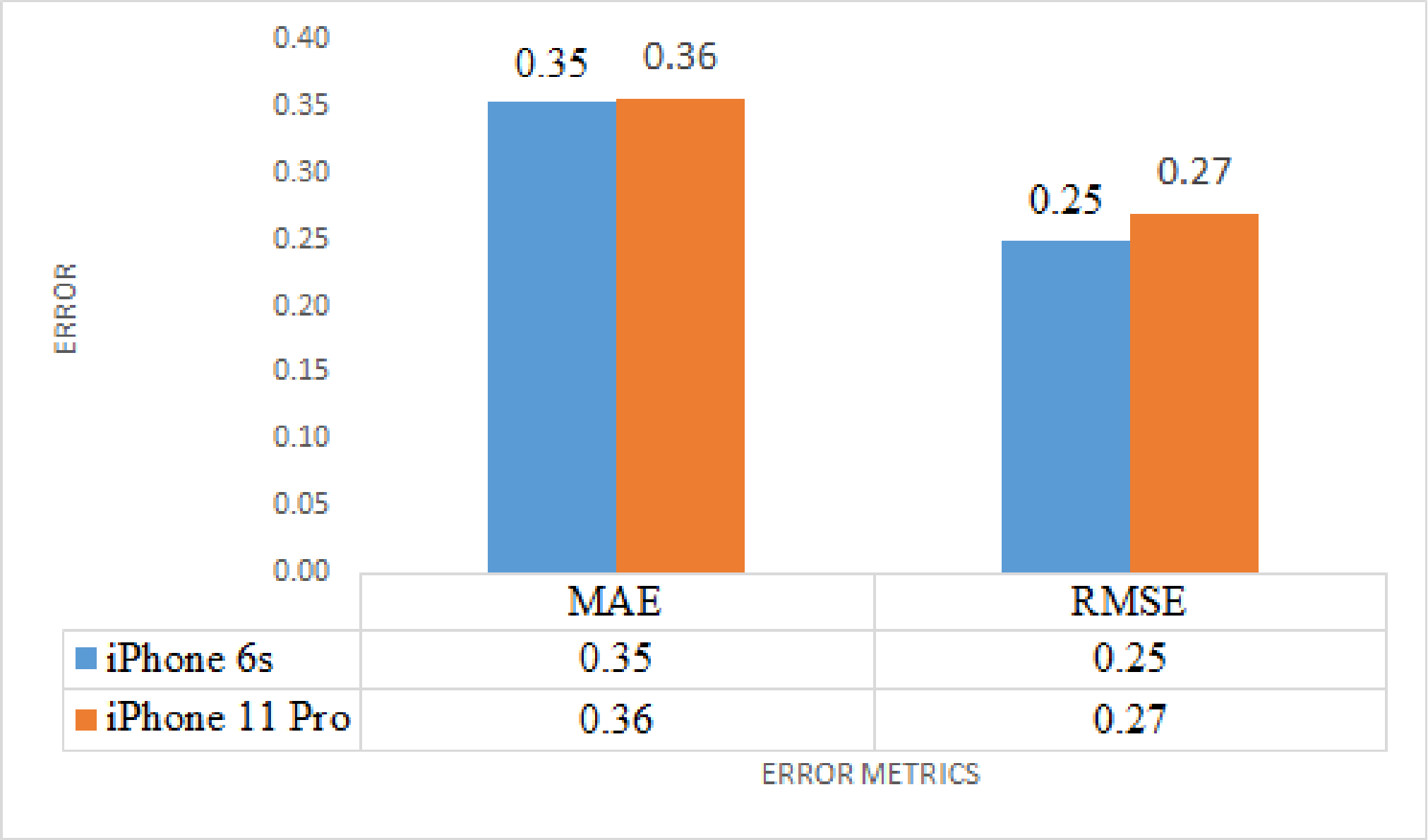

**Figure 10.** Comparison of the performance between the iPhone 6s and iPhone 11 Pro in the SVR model.

The results reveal that predictors of the MLR method yield better results on individual datasets than the SVR and CNN methods for prediction during the holdout cross-validation. However, SVR is better on separate datasets than the other models during k-fold cross-validation and leave-one-out cross-validation. Conversely, weaker predictions are produced when a trained dataset is tested with the other three datasets because data from the trained dataset were not taken under the same sunlight conditions and the identical smartphone versions. In this formation, SVR scores better for datasets 01, 02, and 03 when a test is performed in combination with the three other datasets. Still, MLR and SVR both obtain better results on dataset 04 when the test is executed combined with other datasets. After combining all datasets, SVR achieved better results in k-fold and leave-one-out cross-validation. Still, MLR and SVR both obtain better results in holdout cross-validation. In this research, a single dataset showed better accuracy than the combination because soil images are not taken in the same sunlight conditions (direct or indirect sunlight). Therefore, this study identified that SVR is a better regression model for predicting soil moisture. One reason may be that SVR has a structural risk minimization principle, which minimizes an upper limit of the generalization error rather than minimizing the training error. In contrast, CNN possesses a predefined structure directed toward minimizing the error in the training data [11,39–41]. Secondly, SVR can give more reliable predictions than MLR because SVR considers all linear and nonlinear useful information, whereas MLR only considers the linear relationship between the actual and predicted values [42,43]. Another reason might be the SVR method demonstrates the highest consistency in a small amount of sample prediction and obtains an optimal overall solution, avoiding the local extremity issue that the CNN is subject to. Moreover, the SVR regression model performs better with fewer parameters than the CNN model.

## 5. Conclusions

This paper employed MLR, SVR, and CNN models for predicting soil moisture. Several cross-validation techniques were implemented to understand the variation of soil moisture accuracy on the ML models. SVR achieved better results than the others. This is explained by the fact that the SVR method demonstrated excellent consistency for moisture prediction with a small number of samples. This research also found that direct or indirect sunlight was a significant factor during the capture of the soil images for soil moisture estimation. In this case, indirect sunlight exhibited a better performance in estimating soil moisture. A further finding was that the different smartphone types did not result in a significant distinction in evaluating soil moisture. Overall, this research again demonstrated that a smartphone could be useful for soil moisture estimation, and farmers might benefit from this strategy. Because of the availability of smartphones among farmers in urban or rural areas, this proposed system will be robust for moisture prediction in any agriculture industry. Although certain models performed better using a small dataset in this research, complementary studies with a large dataset are still required to understand the models' performances. Moreover, a better result could be executed by including additional input parameters with soil images during the training of models. In addition, to determine a better ML model, other ML models are needed in the future. Furthermore, this research only focused on multiple versions of the iPhone belonging to one mobile company. Further study is required to determine the effects on test performance with various versions of smartphones from other companies.

**Author Contributions:** Conceptualization, M.R.H.H. and M.A.K.; methodology, M.R.H.H. and M.A.K.; software, M.R.H.H.; validation, M.R.H.H. and M.A.K.; formal analysis, M.R.H.H. and M.A.K.; investigation, M.R.H.H.; resources, M.R.H.H.; data curation, M.R.H.H.; writing—original draft preparation, M.R.H.H.; writing—review and editing, M.A.K.; visualization, M.R.H.H. and M.A.K.; supervision, M.A.K.; project administration, M.A.K. All authors have read and agreed to the published version of the manuscript.

**Funding:** This research received no external funding.

**Institutional Review Board Statement:** Not applicable.

**Informed Consent Statement:** Not applicable.

**Data Availability Statement:** The data presented in this study are available on request from the corresponding author.

**Conflicts of Interest:** The authors declare no conflict of interest.

## Abbreviations

| | | | |
|---|---|---|---|
| ANN | artificial neural network | MLR | multiple linear regression |
| CNN | convolutional neural network | MSE | mean squared error |
| DL | deep learning | OLS | ordinary least squares |
| FDR | frequency domain reflectometry | PLS | partial least squares |
| GPR | Gaussian process regression | RF | random forest |
| GPS | global positioning system | RMSE | root mean square error |
| LCCC | Lin's concordance correlation coefficient | RPD | ratio of performance to deviation |
| LM | light meter | RPIQ | ratio of performance to interquartile distance |
| LR | linear regression | SM | soil moisture |
| MAE | mean absolute error | SVM | support vector machine |
| MLP | multilayer perceptron | SVR | support vector regression |
| ML | machine learning | TDR | time domain reflectometry |